\documentclass[10pt,twocolumn,letterpaper]{article}
\usepackage[pagenumbers]{cvpr} 

\usepackage{graphicx}
\usepackage{amsmath}
\usepackage{amssymb}
\usepackage{booktabs}
\usepackage[toc,page]{appendix}

\usepackage{array}
\usepackage{multirow}
\usepackage{pifont}
\usepackage{amssymb}
\usepackage{subfloat}
\usepackage{algorithm}
\usepackage{algorithmic}
\usepackage{amsfonts}
\usepackage{amsmath}
\usepackage{wrapfig}
\usepackage{graphicx}
\usepackage{booktabs}

\usepackage{multirow}
\usepackage{pifont}

\usepackage{setspace}

\newcommand{\cmark}{\ding{51}}%
\newcommand{\xmark}{\ding{55}}%
\newcommand*{\affaddr}[1]{#1} 

\newcommand*{\email}[1]{\texttt{#1}}
\usepackage[pagebackref,breaklinks,colorlinks]{hyperref}

\usepackage[capitalize]{cleveref}
\crefname{section}{Sec.}{Secs.}
\Crefname{section}{Section}{Sections}
\Crefname{table}{Table}{Tables}
\crefname{table}{Tab.}{Tabs.}

\begin{document}

\title{The Devil is in the Frequency: Geminated Gestalt Autoencoder \\ for Self-Supervised Visual Pre-Training}

\author{%
	Hao Liu\textsuperscript{\dag\thanks{Equal contribution. \textsuperscript{\dag}Contact person.}} \quad Xinghua Jiang\textsuperscript{*}  \quad Xin Li \quad Antai Guo \quad Deqiang Jiang \quad Bo Ren\\		
	\affaddr{Tencent YouTu Lab} \quad\\
	\email{\small \{ivanhliu, clarkjiang, fujikoli, ankerguo, dqiangjiang, timren\}@tencent.com}
}
\maketitle

\begin{abstract}
	The self-supervised Masked Image Modeling~(MIM) schema, following ``mask-and-reconstruct'' pipeline of recovering contents from masked image, has recently captured the increasing interest in the community, owing to the excellent ability of learning visual representation from unlabeled data. Aiming at learning representations with high semantics abstracted, a group of works attempts to reconstruct non-semantic pixels with large-ratio masking strategy, which may suffer from ``over-smoothing'' problem,   while others directly infuse semantics into targets in off-line way requiring extra data. Different from them, we shift the perspective to the Fourier domain which naturally has global perspective and present a new Masked Image Modeling~(MIM), termed Geminated Gestalt Autoencoder~($\text{Ge}^2$-AE) for visual pre-training. Specifically, we equip our model with geminated decoders in charge of reconstructing image contents from both pixel and frequency space, where each other serves as not only the complementation but also the reciprocal constraints. Through this way, more robust representations can be learned in the pre-trained encoders, of which the effectiveness is confirmed by the juxtaposing experimental results on downstream recognition tasks. We also conduct several quantitative and qualitative experiments to investigate the learning behavior of our method. To our best knowledge, this is the first MIM work to solve the visual pre-training through the lens of frequency domain. 
\end{abstract}

\section{Introduction}

\begin{figure}[ht]
	\begin{center}
		\includegraphics[width=1\linewidth]{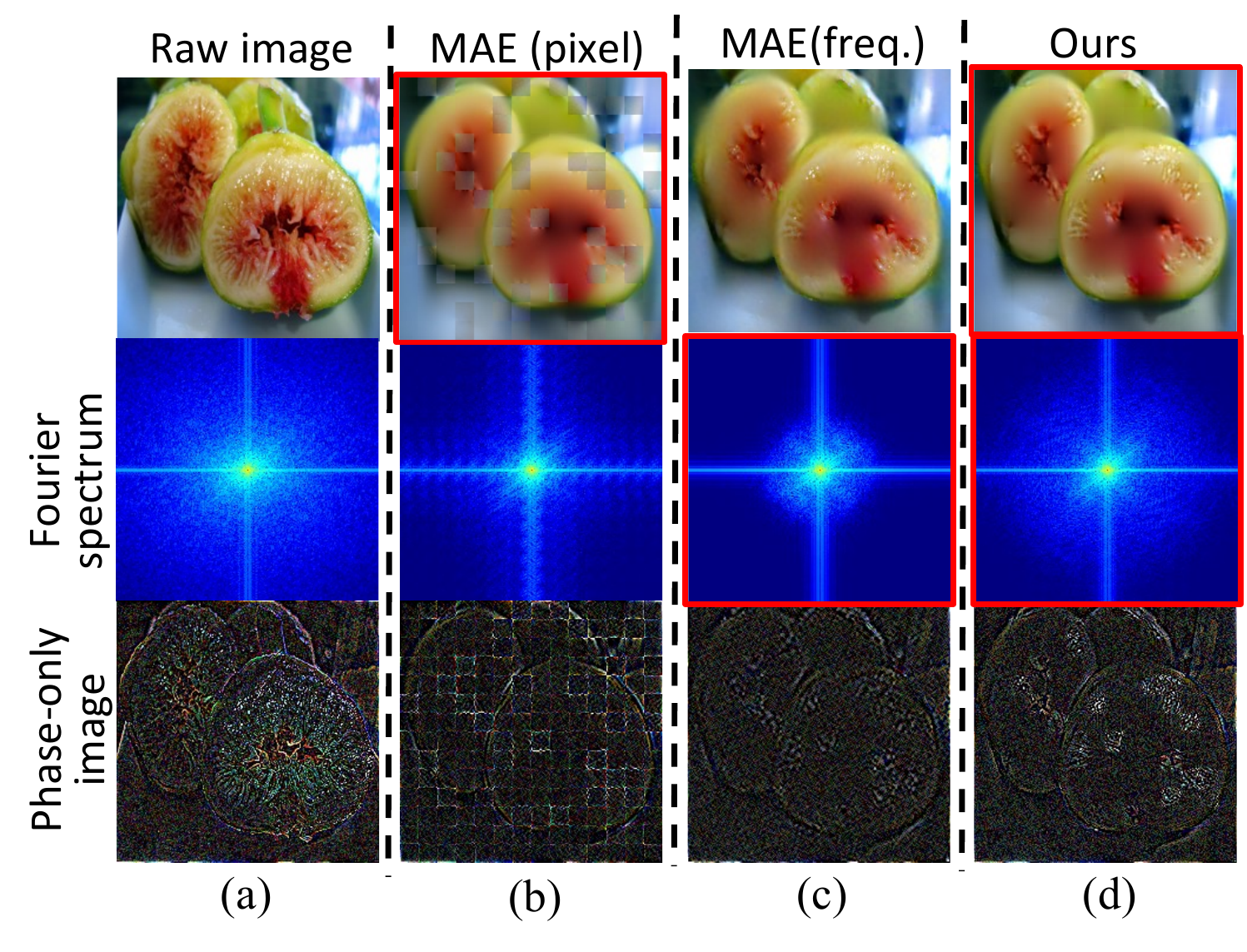}
		\\
		\vspace{-5mm}
	\end{center}
	\caption{Demonstrations of reconstructed images~(1st. row), Fourier spectrum maps~(2nd. row) and phase-only images~(3rd. row) yielded by MAE and our proposed method. The (a) column is the raw image and corresponding maps, while (b) and (c) are results of MAE regarding pixel and frequency as targets. (d) are ours. The direct predictions are highlighted by red boundaries, and their corresponding pixel or frequency maps are obtained by 2D-FFT or 2D-IFFT. The phase-only images indicating semantics are obtained by setting the amplitude component to a constant. 
	}\label{fig:moti}
	\vspace{-3mm}
\end{figure}

Recently, self-supervised visual pre-training has witnessed a fast-paced progress in learning robust representations of visual content, which can overcome the \textit{data appetite} problem suffered by supervised learning paradigm, where large amount of labeled training data is required. Among, a series of methods, termed Masked Image Modeling~(MIM)~\cite{he2021masked, wei2021masked, xie2021simmim,dong2021peco,zhou2021ibot,chen2022context}, exhibits promising potential, which inherits the ``mask-and-reconstruct'' thought from masked autoencoding methods in natural language processing~(NLP) field, such as BERT\cite{devlin2018bert}. More concretely, parts of content in input image are masked to learn latent representations from the visible regions by encoder, which are then used to reconstruct content by decoder. For the downstream visual tasks, the encoder is expected to yield robust representation with \textit{rich semantic information}. However, the language naturally has semantics highly abstracted by human~\cite{he2021masked}, where the elemental unit is also discrete. In contrast, pixels contained in image merely contain low-level statistics and often present the heavy redundancy due to the continuous property of image. Therefore, the targets setting become the crucial component in the MIM, which essentially determine what unwanted minutiae should be compressed for the semantic-perceiving purpose.    

To implement the purpose, existing MIM methods have made attempts mainly in two directions \textit{w.r.t} the target setting. Among, MAE~\cite{he2021masked} and SimMIM~\cite{xie2021simmim} are two simple and straightforward methods, which regard raw pixels as targets and encourage the model concentrating on semantics by large-ratio masking strategy. Although ViT~\cite{dosovitskiy2020image} is adopted, the target pixels serving as non-semantic entities still contain heavy redundancy and lack global perspective, which is essential to the high semantics capture. Alternatively, extra knowledge is introduced in targets to achieve the goal in tokenizer-based methods. Nevertheless, there is no free lunch. These methods reconstruct content supervised by off-line or on-line trained tokenizers~\cite{bao2021beit, zhou2021ibot, dong2021peco}, which may inevitably incur complicated processing, such as knowledge distillation or contrastive learning~\cite{zhou2021ibot}. Although MaskFeat~\cite{wei2021masked} simply regards handcrafted HOG feature as target~\cite{wei2021masked}, the target is still a local statistic-based descriptor, which may also suffer from the same problem with pixel prediction-based methods. Conclusively, a qualified supervision should present the following properties: 1)~\textit{well discretizing continuous image content}; 2)~\textit{containing high visual semantic}. We ask: ``\textit{Is there a simple way to obtain the supervision with the desired properties merely from image \textit{per se}?}''  

To answer this question, we shift our perspective to the frequency domain. Our motivation comes from several intriguing properties of the Fourier transformation. Firstly, as suggested by many works~\cite{oppenheim1979phase, oppenheim1981importance, piotrowski1982demonstration, hansen2007structural}, high-level semantics of the original signal are naturally preserved in the phase component of Fourier spectrum, while the amplitude component contains low-level statistics. It can be vividly demonstrated by the clear object contour in the phase-only image~(last image of Fig.~\ref{fig:moti}(a)), where the amplitude component is set to a constant. Besides, Fourier basis are orthogonal, which can spontaneously discretize spatial domain content into individual frequency component in Fourier spectrum. Last but not least, each frequency component in Fourier spectrum has the global vision, which of each is also, to put it another way, a high summary of whole image. 

By visualizing the Fourier spectrum of recovered images of MAE~(Fig.~\ref{fig:moti}(b)), we find that a portion of high frequency is compressed while low frequency is dominant. Although the phase-only image exhibits semantic contour to certain extent, there are also some unwanted details preserved, \textit{e.g., grid like texture caused by independent pixel reconstruction process}. As a result, the integrity of object contour would be disrupted. Alternatively, we come up with a solution by directly replacing the pixel target in MAE~\cite{he2021masked} with the Fourier spectrum where each component carries the global information. From the phase-only image of Fig.~\ref{fig:moti}(c), although a more holistic contour can be observed, the pre-trained encoder still can not yield decent representation on downstream tasks. We attribute it to the ``over-smoothing'' issue, \textit{e.g.}, too many useful information hidden in high or middle frequencies are abandoned, as illustrated by Fourier spectrum in Fig.~\ref{fig:moti}(c). That is to say, the model overly pays  attention to semantics while certain local details are ignored, whereas a good representation needs both semantic and local details, as suggested by work~\cite{dong2021peco}. To achieve this goal and substantially retain the merit of ``simple yet effective'', we build our MIM method, termed \textbf{Ge}minated \textbf{Ge}stalt \textbf{A}uto\textbf{E}ncoder~($\text{Ge}^2$-AE), upon canonical MAE~\cite{he2021masked} and simply modify it with one extra lightweight frequency decoder~(FD) added to simultaneously perform gestalt tasks of the local masked region and global frequency. To mitigate the inconsistency between pixel and frequency space, the task-specific FD is equipped with Fourier Spectrum Perceiver~(FSP) and process spatial-frequency contents alternatively to adapt the spatial encoded tokens to the frequency prediction task. Moreover, the information in pixel and frequency space also play as the reciprocal constraints dressed on each other when reconstructing contents from both domains. As illustrated by Fig.~\ref{fig:moti}(d), despite that the reconstructed image of $\text{Ge}^2$-AE has no obvious difference from MAE, the yielded Fourier spectrum overcomes the ``over-smoothing'' problem with more proper details and holistic object contour~(refer to phase-only image in Fig.~\ref{fig:moti}(d)) preserved. Benefiting from the tailored design, our $\text{Ge}^2$-AE can achieve significant performance improvement on pre-trained representation quality than other methods, and shows inspiring transferability, which are validated by the experimental results on several downstream visual recognition tasks. 
In summary, our contributions are:        
\begin{itemize}
	\item We reinspect the Masked Image Modeling for visual pre-training task from frequency domain and investigate intriguing properties for robust representation learning. To our best knowledge, we are the first to explore the space domain imperceptible clues from frequency perspective in visual pre-training field.
	
	\item We coin a novel $\text{Ge}^2$-AE tailored for visual pre-training problem, which consists of geminated decoders performing gestalt from both spatial and frequency domain, which can fully leverage their reciprocity to enhance the semantics without introducing extra complicated processes. 
	
	\item Quantitative and qualitative experimental results on public benchmarks demonstrate that our method achieve significant performance improvement on pre-trained representation quality than state-of-the-arts.
	
\end{itemize}

\section{Related Work}

\subsection{Visual Pre-training}
Recent self-supervised visual pre-training techniques can be roughly categorized into two types: \textit{contrastive learning} and \textit{Masked Image Modeling}. We briefly review them as follows:

\textbf{Contrastive learning.}\quad In the early stage, the contrastive learning methods with contrastive loss~\cite{he2020momentum} as \textit{de facto} component have heavily boosted the transfer performance of visual models and thus become the mainstream in the self-supervised visual pre-training field. Among, He et al.~\cite{he2020momentum} propose a momentum contrast-based method to overcome the  batch size limitation, bringing transfer performance of pre-training models a huge leap forward. As a concurrent work, the SimCLR~\cite{chen2020simple} investigates the importance of various data augmentations in contrastive learning. To avoid the sampled negative noise problem, BYOL~\cite{grill2020bootstrap} proposes to simply push the positive pairs together. As a powerful backbone, vision transformer~(ViT)~\cite{dosovitskiy2020image} and its variants~\cite{wang2021pyramid, liu2021swin, liu2021nommer, wu2021cvt} have shown superior performance in many visual tasks than canonical convolutional networks. For ViTs, several self-supervised methods~\cite{caron2021emerging, chen2021empirical} aiming at more effectively pre-training them are proposed. 

\textbf{Masked Image Modeling.}\quad As the upstart in self-supervised visual pre-training, Masked Image Modeling~(MIM)-based approaches have been proposed recently and exhibit promising potentials, which are mostly inspired by the success of Masked Language Modeling~(MLM) in NLP field. Following the ``mask-and-reconstruct'' pipelines, existing MIM methods design the pretext task with either pixel-level information reconstruction or token prediction. 

On the pixel-level aspect, MAE~\cite{he2021masked} and SimMIM simply mask the pixels in the patches and then predict them to encourage model focus on the semantics. Similarly, CiM~\cite{fang2022corrupted} and CAE~\cite{chen2022context} are proposed to achieve the same goal but with more sophisticated structure designs. However, some useful local details may lost in these methods. Although MaskFeat~\cite{wei2021masked} introduces the low-level local descriptors, HOG, as targets, the problem still exists.

In the direction of token-level prediction, BEiT~\cite{bao2021beit} as a pioneer work concludes images into the off-line trained tokens as pre-training targets, which aims to integrate semantics into models through discretized tokens. Concurrently, PeCo~\cite{dong2021peco} introduces the thought of contrastive learning to learn a perceptual visual vocabulary for pre-training. In the similar way, iBOT~\cite{zhou2021ibot} marries MIM with knowledge distillation to obtain the tokenizer in on-line manner. Although the tokenizer-based methods can introduce more semantics to the model, there may require extra data, \textit{e.g.}, DALLE~\cite{ramesh2021zero} or incur complicated data augmentation along with the contrastive learning. In contrast to above methods, our proposed method can achieve the similar semantic-perceiving purpose but without extra data or contrastive operation.   

\subsection{Learning in Frequency Domain}
Considering the exclusive properties, more and more researches transform image to frequency domain for the subsequent processing. Based on Discrete Cosine Transform~(DCT), DCTransformer~\cite{nash2021generating} is proposed to sequentially generate high resolution images from sparse representations obtained by DCT. Inspired by the working mechanism of human visual system where different frequency components are treated unequally, method~\cite{xu2020learning} inputs the image preprocessed by DCT into prevalent deep model to learn visual representations while work~\cite{liu2021nommer} regards DCT-quantized features as a candidate representing rough global information to be selected in the learning procedure. In addition, inspired by many early investigations~\cite{oppenheim1979phase, oppenheim1981importance, piotrowski1982demonstration, hansen2007structural} on signal from Fourier domain, the method~\cite{xu2021fourier} deploys Fourier transform to achieve the semantic enhancement. Similarly, the work~\cite{jiang2021focal} designs a frequency loss to facilitate optimization in the frequency dimension, driving frequencies of an image can be well represented and distinguished. Furthermore, there also exist several investigative literatures~\cite{rahaman2019spectral, tancik2020fourier, wang2020high} to explore the learning bias from frequency perspective. In contrast to previous methods, to our best knowledge, $\text{Ge}^2$-AE is the first work to study the effectiveness of introducing Fourier analysis to MIM for visual pre-training through the lens of frequency domain.  
\section{The Methodology}
\subsection{Preliminary: 2D-Discrete Fourier Transform }
Before introducing our method, we first briefly review the 2D-Discrete Fourier Transform~(DFT) serving as an important tool in signal analysis, which plays an indispensable role in our $\text{Ge}^2$-AE. Given a 2D signal~(one channel of image or feature cube) $\mathbf{F} \in \mathbb{R}^{W \times H}$, its  2D-DFT can be defined as:
\begin{align}
	\small
	f(u, v)=\sum_{h=0}^{H-1} \sum_{w=0}^{W-1} F{(h, w)} e^{-j 2 \pi\left(\frac{uh}{H}+\frac{vw}{W}\right)},\label{eq:fft}
\end{align}
where $F{(h, w)}$ represents the $h$-th and $w$-th pixel or element in $\mathbf{F}$ while the $u$ and $v$ are indexes of horizontal and vertical spatial frequencies in Fourier spectrum. Correspondingly, the 2D Inverse DFT~(2D-IDFT) is formulated as:  
\begin{align}
	\small
	F(h, w)=\frac{1}{H W} \sum_{u=0}^{H-1} \sum_{v=0}^{W-1} f(u, v) e^{j 2 \pi\left(\frac{uh}{H}+\frac{vw}{W}\right)},\label{eq:ifft}
\end{align}
Both DFT and IDFT can be accelerated with the fast version, FFT algorithm~\cite{nussbaumer1981fast}. Moreover, the definitions of amplitude $A$ and phase components $P$ deduced from Fourier spectrum are given as: 
\begin{align}
	\small
	A(u, v) &=\left[\mathcal{R}^{2}(u, v)+\mathcal{I}^{2}(u, v)\right]^{1 / 2}, \\
	P(u, v) &=\arctan \left[\frac{\mathcal{I}(u, v)}{\mathcal{R}(u, v)}\right],
\end{align}
where $\mathcal{R}$ and $\mathcal{I}$ denote the real and imaginary part of frequency $f$ respectively, \textit{i.e.}, $f(u,v) = \mathcal{R}(u, v)+\mathcal{I}(u, v)i$.
\subsection{Architecture of $\text{Ge}^2$-AE}
\begin{figure*}[ht]
	\begin{center}
		\includegraphics[width=0.92\linewidth,height=0.35\linewidth]{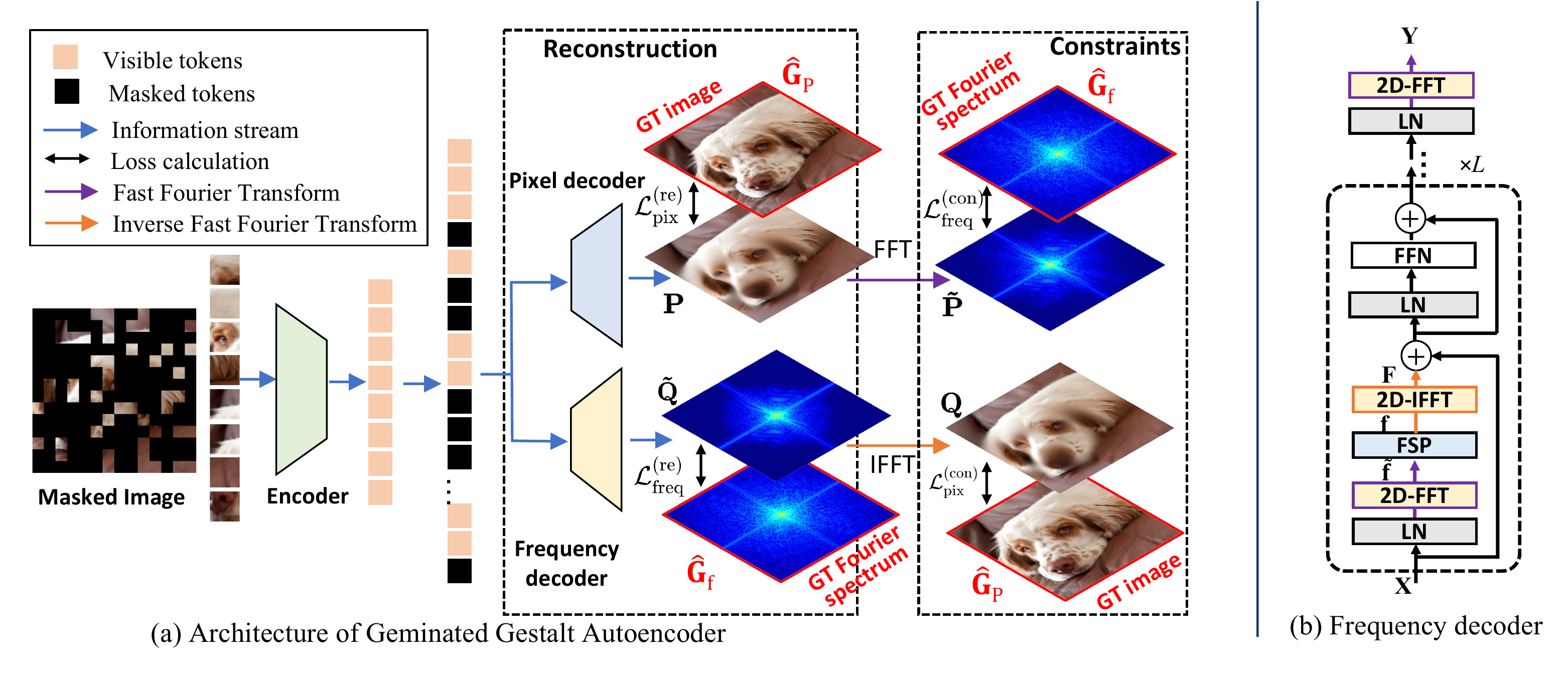}
		\\
		\vspace{-6mm}
	\end{center} 
	\caption{Architecture of the Proposed Geminated Gestalt Autoencoder~($\text{Ge}^2$-AE). The encoder receives the unmasked patches to yield visible tokens, which are sent to the geminated structure decoders together with masked tokens to recover in pixel and Fourier domain constrained by each other. Best viewed in color and zoomed in.
	}\label{fig:arch}
	\vspace{-3mm}
\end{figure*}
The architecture of our proposed $\text{Ge}^2$-AE is illustrated in Fig.~\ref{fig:arch}(a), which is built upon a representative baseline, Masked Autoencoder~(MAE). Specifically, we inherit the encoder design of MAE, where the canonical ViT is adopted to project the unmasked patches into visible tokens. As suggested by MAE~\cite{he2021masked}, the decoder design is crucial to the MIM model, as it not only learns representations of masked tokens,  but also determines the semantic level of the whole learned latent representations. 

The pre-training task-tailored designs of our method lie in the following three folds: \textbf{I)~Geminated structure}. Considering the raw pixels regarded as targets in MAE are non-semantic entities, we simply replace pixel targets with Fourier spectrum containing global semantics. However, we observe the evident performance drops. Therefore, we equip our MIM pre-training framework with decoupled decoders in charge of reconstructing spatial and frequency domain simultaneously. The assumption behind our design is that \textit{a good latent representation should be a consensus between local details and global semantics.} We preserve the decoder from the vanilla MAE as the pixel decoder~(PD) for the similar purpose, which deploys ViT blocks as the core ingredients. \textbf{II)~Frequency decoder}. As for the frequency decoder~(FD), we modify the vanilla ViT blocks to adapt to the frequency gestalt task, which will be elaborated in Sec.~\ref{sec:fq}. \textbf{III)~Complementary constraints.} The information of spatial and frequency domain are not only expected to be predicted, but also can provide the complementary constraints for each other. As shown in Fig.~\ref{fig:arch}(a), taking the pixel reconstruction branch for example, we further convert the predicted image to Fourier domain by FFT and treat the Fourier spectrum of raw image as constraint for it. To put it in another way, this constraint encourages the global frequency involving semantics preserved at most when reconstructing image. Correspondingly, the pixel constraint put on reconstructed Fourier spectrum aims to preserve more local content details in the predicted spectrum.

\subsection{Frequency Decoder}\label{sec:fq}
As described above, the frequency decoder~(FD) in our method plays a role in infusing the global frequency information to the learned latent representation. A plain way to implement it is to apply FFT either on the input or the output of decoder. However, neither the early-FFT nor late-FFT process can well adapt to the task of directly predicting global frequency from spatial encoded tokens, not to mention that some of them are masked. Alternatively, we consider to achieve the above purpose in a progressive way to alleviate the inconsistency problem of spatial-frequency domain. Inspired by the frequency usage in NomMer~\cite{liu2021nommer}, as demonstrated in Fig.~\ref{fig:arch}(b), we make several task-specific modifications on the vanilla ViT blocks. Specifically, the basic block takes spatial features as input appended by a Layer Normalization~(LN)~\cite{vaswani2017attention}, then it is processed by the 2D-FFT~(purple arrow) defined in Eqn.~\eqref{eq:fft}. Once converted to the frequency spectrum $\mathbf{\tilde{f}} \in \mathbb{R}^{W \times H \times C}$, each component in it can have global vision on the whole image but comprised by different vision patterns, \textit{e.g.,} texture of dog hairs. To highlight those significant frequency components, which could be propagated to the learned latent representations and informative for visual recognition, we insert a Fourier Spectrum Perceiver~(FSP) inside FD, formulated as $\mathbf{f} = \mathbf{\Omega} \odot \mathbf{\tilde{f}}$, where $\mathbf{\Omega} \in \mathbb{R}^{W \times H \times C}$ is the learnable parameter matrix. ``$\odot$'' represents the Hadamard product. Once the emphasized frequency $\mathbf{f}$ obtained, the 2D-IFFT~(orange arrow) defined in Eqn.~\eqref{eq:ifft} is applied to restore the spatial feature $\mathbf{F}$. Afterwards, Feed-Forward Networks~(FFN)~\cite{vaswani2017attention} equipped with residual connection and LN are appended, which is similar to that of vanilla ViT block. These basic blocks can be stacked $L$ times, enabling the alternative reconstruction of spatial-frequency contents, which can feasibly solve the inconsistency problem. In the end, the 2D-FFT with LN is applied to predict the final reconstructed Fourier spectrum.

Moreover, it is worth mentioning that, compared to vanilla ViT blocks deploying self-attention operation with $\mathcal{O}(n^2)$ complexity~($n$ is the token number), our FD is a lightweight module. In detail, the total computational complexity of FD's basic block is $\mathcal{O}(n\log n)$, where 2D-FFT and 2D-IFFT are both with $\mathcal{O}(n\log n)$ complexity while that of Hadamard product is only $\mathcal{O}(n)$.  

\subsection{Pre-training Strategy}
\textbf{Overall Loss.}\quad During pre-training, our $\text{Ge}^2$-AE learns latent representation by solving content gestalt from both pixel-level and frequency-level:
\begin{align}
	\small
	\mathcal{L} &= \overbrace{\mathcal{L}^{(\text{re})}_{\text{pix}}(\mathbf{P}, \mathbf{\hat{G}}_{\text{p}})+ \mathcal{L}^{(\text{con})}_{\text{freq}}(\mathbf{\tilde{P}}, \mathbf{\hat{G}}_{\text{f}})}^{pixel-level }\\\notag
	&+  \underbrace{\lambda (\mathcal{L}^{(\text{re})}_{\text{freq}}(\mathbf{\tilde{Q}}, \mathbf{\hat{G}}_{\text{f}})
		+ \mathcal{L}^{(\text{con})}_{\text{pix}}(\mathbf{Q}, \mathbf{\hat{G}}_{\text{p}}))}_{frequency-level},
	\label{overall-loss}
\end{align}
where $\mathcal{L}^{(re)}_{\sim}$ represent losses for reconstruction task while $\mathcal{L}^{(con)}_{\sim}$ 
serve as constraints, as illustrated in Fig.~\ref{fig:arch}(a). $\mathbf{P}$ and $\mathbf{\tilde{Q}}$ are the predictions of pixel and frequency decoders while $\mathbf{\tilde{P}}$ and $\mathbf{Q}$ are their corresponding transformations obtained by 2D-FFT and 2D-IFFT. $\mathbf{\hat{G}}_{\text{p}}$ and $\mathbf{\hat{G}}_{\text{f}}$ are the raw image and its Fourier spectrum serving as Ground Truths~(GTs) for both reconstruction and subsequent constraints. The effect of either decoder is controlled by loss weight $\lambda$ ($\lambda$ =0.5 in default). For the pixel loss $\mathcal{L}^{\sim}_{\text{pix}}$, we compute Mean Square Error~(MSE) between the reconstructed and raw images in pixel space, which is similar to MAE~\cite{he2021masked}.    

\textbf{Frequency loss.}\quad As studied in previous works~\cite{rahaman2019spectral, xu2022overview, wang2020high} on the learning behavior from frequency domain, \textit{spectral bias} of the deep neural networks is often inclined to low frequency functions. Besides, according to \textit{F-Principle}~\cite{xu2019frequency}, the fitting priority of a network to certain frequencies is various throughout the training, often in the low-to-high pattern. As a result, for our frequency decoder, some significant frequency could be hardly decoded, once those frequencies with higher priority are generated. Specifically, high-priority frequency is termed \textit{easy frequency}, otherwise \textit{hard frequency.} 

To better capture the hard frequency, we adopt focal frequency loss~\cite{jiang2021focal} as $\mathcal{L}_{\text{freq}}$ to dynamically tune each frequency weight, which can be defined as:

\begin{align}
	\small
	&\mathcal{L}_{\text{freq}} =\frac{1}{H W} \sum_{u=0}^{H-1} \sum_{v=0}^{W-1} \omega(u, v)\odot \gamma(f(u, v),\hat{f}(u, v))^2, 
\end{align}	
where $f(u,v)$ is the $i, j$-th frequency component of spectrum $\tilde{\mathbf{Q}}$ or $\tilde{\mathbf{P}}$ while $\hat{f}(u,v)$ denotes the frequency component from GT spectrum $\hat{\mathbf{G}}_{\text{freq}}$ at the same location. In addition, $\gamma(f,\hat{f})$ is the frequency distance implemented by computing squared Euclidean distance between their real and imaginary parts. $\omega$ is spectrum weight matrix can down-weight the easy frequencies, which are formulated as: 
\begin{align}
	\small
	&\omega(u, v)=\gamma(f(u, v),\hat{f}(u, v))^{\beta},\\
	&\gamma(f,\hat{f}) = \sqrt{(\mathcal{R}- \tilde{\mathcal{R}})^2+(\mathcal{I}- \tilde{\mathcal{I}})^2 },
\end{align}
$\beta$ is the scaling factor for flexibility, which is set to 1 in default.

\section{Experiments}
\subsection{Implementation Details}
For the pre-training on ImageNet-1K~(IN1K) training set, we inherit the experimental settings in MAE~\cite{he2021masked}. The random mask ratio is set to 75\% in default while the block number $L$ is empirically set to 8 for both PD and FD. Moreover, with the initial learning rate and batch size are set to $1.5e^{-4}$ and 4,096, respectively, all our models are pre-trained for 800 epochs with the input size of $224^2$. Besides, AdamW~\cite{loshchilov2017decoupled} optimizer with a cosine learning rate scheduler is adopted. We implement the $\text{Ge}^2$-AE architecture by Pytorch~\cite{paszke2019pytorch} with all experiments conducted on a workstation with 32
NVIDIA A100-40 GB GPUs. All the reported results are the averaged ones over 10 random seeds.
Please refer to appendix for more details.

\subsection{Image Classification on ImageNet-1K}
\begin{table}
	\centering
	\small
	\setlength{\tabcolsep}{3mm}
	\caption{ImageNet-1K (IN1K) fine-tuning Top-1 accuracy of ViTs with different sizes.}
	\label{tab:finetune}
	\begin{tabular}{l|cccc}
		\toprule
		Method & ViT-S & ViT-B & ViT-L & ViT-H\\
		\midrule
		DINO~\cite{caron2021emerging} & - & 82.8 & - & -\\
		CAE~\cite{chen2022context}	& 81.8 & 83.6 & - & -\\
		CIM~\cite{fang2022corrupted} & 81.6 & 83.1 & - & -\\
		BEiT~\cite{bao2021beit} & - & 83.2 & 85.2 & -\\
		MoCo v3~\cite{chen2021empirical} & - & 83.2 & 84.1 & -\\
		MaskFeat~\cite{wei2021masked} & - & 84.0 & 85.7 & -\\
		iBOT~\cite{zhou2021ibot} & \textbf{82.3} & 84.0 & 85.2 & -\\
		PeCo~\cite{dong2021peco} & - & 84.5 & 86.5 & 87.5\\
		SimMIM~\cite{xie2021simmim} & - & 83.8 & - & -\\
		MAE~\cite{he2021masked} & - & 83.6 & 85.9 & 86.9\\
		\midrule
		\textbf{$\text{Ge}^2$-AE~(Ours)} & 82.2 & \textbf{84.8} & \textbf{86.6} & \textbf{87.7}	\\		
		\bottomrule
	\end{tabular}
\end{table}

\begin{table}
	\centering
	\small
	\setlength{\tabcolsep}{2.5mm}
	\caption{ImageNet-1K~(IN1K) Top-1 accuracy of different methods under linear probing setting.}
	\label{tab:knn&linear}
	\begin{tabular}{l|c|c|c|c}
		\toprule
		Method & Arch. & Param. & Epoch& Linear\\
		\midrule
		DINO~\cite{caron2021emerging} & ViT-S & 21 & - & 77.0\\
		DINO~\cite{caron2021emerging} & ViT-B & 85 & - & 78.2\\
		MoCo v3~\cite{chen2021empirical} & ViT-B & 86 & -  & 76.7\\
		CAE~\cite{chen2022context} & ViT-S & 21 & 300  & 50.8\\
		CAE~\cite{chen2022context} & ViT-B & 85 & 800  & 68.3\\
		iBOT~\cite{zhou2021ibot} & ViT-S & 21 & 3,200  & 77.9\\
		iBOT~\cite{zhou2021ibot} & ViT-B & 85 & 1,600 & \textbf{79.5}\\
		SimMIM~\cite{xie2021simmim} & ViT-B & 85 & -  & 56.7\\
		MAE~\cite{he2021masked} & ViT-B & 86 & 800& 68\\
		\midrule
		\textbf{$\text{Ge}^2$-AE~(Ours) } & ViT-B & 86 & 800  & 75.3\\
		\bottomrule
	\end{tabular}
\end{table}
We firstly perform self-supervised pre-training on the ImageNet-1K~(IN1K)~\cite{deng2009imagenet} training set and evaluate the learned representation quality with only the encoder preserved under two supervised training settings: \textbf{1)~end-to-end fine-tuning; 2)~linear probing}.

\textbf{Experimental setting.}
For the end-to-end fine-tuning, our model employing ViT-base as default encoder is trained for 300 epochs with the input size of $224^2$, of which the initial learning rate and batch size are set to $1e^{-3}$ and 1,024, respectively. Besides, AdamW~\cite{loshchilov2017decoupled} optimizer with a cosine learning rate scheduler is adopted. The weight decay is set to 0.05 and the maximal gradient norm is clipped to 5.0. As for the linear probing setting, we exploit the LARS~\cite{you2017large} optimizer with batch size 16,384 and 0.1 learning rate, and train the model for 100 epochs. Please refer to appendix for more details.

\textbf{End-to-end fine-tuning.}\quad Under this setting, the pre-trained encoder is fine-tuned with the classification head together. From the Tab.~\ref{tab:finetune}, one can observe that our $\text{Ge}^2$-AE can achieve round 1\% performance improvement than the baseline method MAE. Compared to those MIM methods~(PeCo, iBOT) introducing complex contrastive operations yield tokenizers, our method can also achieve the competitive results for ViTs in different model sizes, but with only a lightweight frequency decoder added. We attribute it to the good global semantic perceiving ability of our method, which is comparable to the tokenizer-based methods.

\textbf{Linear probing.}\quad When evaluated under linear probing setting, all parameters of the pre-trained encoder are frozen while only the last classification layer is trained. From Tab.~\ref{tab:knn&linear}, we surprisingly find that our method can surpass the MAE by round 7\% with the similar pre-training epochs. Although the performance of our method is slightly lower than most contrastive learning-based methods~(\textit{e.g.}, DINO, MoCo v3 and iBOT), our method only introduces a lightweight frequency decoder without involving complicated contrastive sample selection or data augmentation. Moreover, this mitigation phenomenon on the gap between MIM and contrastive learning also demonstrates the effectiveness of reconstructing the global semantic information. 

\subsection{Transfer Learning Experiments}
\begin{table}
	\centering
	\small
	\setlength{\tabcolsep}{1mm}
	\caption{Results on COCO object detection and instance segmentation with Mask R-CNN.}
	\label{tab:coco}
	\begin{tabular}{l|l|cccc}
		\toprule
		\multirow{2}{*}{Method}&\multirow{2}{*}{Pre-train data } &\multicolumn{2}{c}{$AP^{box}$} & \multicolumn{2}{c}{$AP^{mask}$}\\
		& & ViT-B & ViT-L & ViT-B & ViT-L\\
		\midrule
		Supervised~\cite{he2021masked} & IN1K w/ labels & 47.9 & 49.3 & 42.9 & 43.9\\
		MoCo v3~\cite{he2021masked} & IN1K & 47.9 & 49.3 & 42.7 & 44.0\\
		BEiT~\cite{he2021masked} & IN1K+DALLE & 49.8 & 53.3 & 44.4 & 47.1\\
		iBOT~\cite{zhou2021ibot} & IN1K & \textbf{51.2} & - & 44.2 & -\\
		PeCo~\cite{dong2021peco} & IN1K & 43.9 & - & 39.8 & -\\
		CAE~\cite{chen2022context} & IN1K & 49.2 & - & 43.3 & -\\
		MAE~\cite{he2021masked} & IN1K & 50.3 & 53.3 & 44.9 & 47.2\\
		\midrule
		\textbf{$\text{Ge}^2$-AE~(Ours)} & IN1K &\text{ 51.0}& \textbf{53.6}& \textbf{45.3} & \textbf{47.8}\\
		\bottomrule
	\end{tabular}
\end{table}

\begin{table}
	\centering
	\small
	\setlength{\tabcolsep}{1mm}
	\caption{Transfer learning accuracy on classification datasets.}
	\label{tab:transfer classification}
	\begin{tabular}{l|cccccc}
		\toprule
		Method & {$Cif_{10}$} & {$Cif_{100}$} & {$iNat_{18}$} & {$iNat_{19}$} & \textit{Flwrs} & \textit{Cars}\\
		\midrule
		\textsl{ViT-S/16}\\					
		DINO~\cite{caron2021emerging} & 99.0 & 90.5 & 72.0 & 78.2 & 98.5 & 93.0\\
		iBOT~\cite{zhou2021ibot} & 99.1 & 90.7 & 73.7 & 78.5 & 98.6 & 94.0\\
		\midrule
		\textbf{$\text{Ge}^2$-AE~(Ours)} & \textbf{99.1 }& \textbf{91.0} &\textbf{74.9}  & \textbf{79.0} & \textbf{98.9} & \textbf{94.5}\\
		\midrule
		\textsl{ViT-B/16}\\			
		DINO~\cite{caron2021emerging} & 99.1 & 91.7 & 72.6 & 78.6 & 98.8 & 93.0\\
		iBOT~\cite{zhou2021ibot} & 99.2 & 92.2 & 74.6 & 79.6 & 98.9 & 94.3\\
		MAE~\cite{he2021masked} & - & - & 75.4 & 80.5 & - & -\\
		\midrule
		\textbf{$\text{Ge}^2$-AE~(Ours)}  & \textbf{99.3 }& \textbf{93.1} & \textbf{75.8} & \textbf{81.2} & \textbf{99.6} & \textbf{94.9}\\
		\bottomrule
	\end{tabular}
\end{table}

\textbf{Object Detection on COCO.}\quad
To verify $\text{Ge}^2$-AE's transferability, we benchmark it on object detection with COCO~\cite{lin2014microsoft}. Following the baseline method MAE~\cite{he2021masked}, the IN1K pre-trained encoder is adopted for initializing the ViT-like backbone of Mask R-CNN~\cite{he2017mask} framework. Besides, FPN~\cite{lin2017feature} is adapted to it~(Refer to MAE~\cite{he2021masked} for more details). We use AdamW~\cite{loshchilov2017decoupled} for optimization with initial learning rate $1.6e^{-4}$ and weight decay 0.1.

As vividly shown in Tab.~\ref{tab:coco}, on this visual prediction task, MIM-based methods~(BEiT, CAE, MAE and our $\text{Ge}^2$-AE) can present more powerful performance than most contrastive learning-based ones~(MoCo v3, PeCo). Although iBOT combining MIM and contrastive learning can achieve the best AP$^\text{box}$, our method can achieve the comparable AP$^\text{box}$ and the best AP$^\text{mask}$ than other compared methods. Especially to the strong baseline MAE, our method can achieve round 0.6\% improvement for either base or large ViT under both protocols, which confirms the necessity of global frequency information in this task.

\textbf{Semantic Segmentation on ADE20K.}\quad
We further evaluate our model on another dense prediction task,
Semantic Segmentation on ADE20K dataset~\cite{zhou2017scene}. Specifically, the encoder pre-trained on IN1K by our $\text{Ge}^2$-AE serves as the backbone of UperNet~\cite{xiao2018unified} which is a prevalent segmentation method. Unless explicitly specified, we use a standard recipe provided by MAE~\cite{he2021masked} to train the models for 100 epochs with batch size 16. 
\begin{table}
	\centering
	\small
	\setlength{\tabcolsep}{3mm}
	\caption{Performance (mIoU) on the ADE20K segmentation task with UperNet framework.}
	\label{tab:ade20k}
	\begin{tabular}{l|l|cc}
		\toprule
		Method & Pre-train data & ViT-B & ViT-L\\
		\midrule
		Supervised~\cite{he2021masked} & IN1K w/ labels & 47.4 & 49.9\\
		MoCo v3~\cite{he2021masked} & IN1K & 47.3 & 49.1\\
		BEiT~\cite{he2021masked} & IN1K+DALLE & 47.1 & 53.3\\
		iBOT~\cite{zhou2021ibot} & IN1K & 50.0 & -\\
		PeCo~\cite{dong2021peco} & IN1K & 46.7 & -\\
		CIM~\cite{fang2022corrupted} & IN1K & 43.5 & -\\
		CAE~\cite{chen2022context} & IN1K & 48.8 & -\\
		MAE~\cite{he2021masked} & IN1K & 48.1 & 53.6\\
		\midrule
		\textbf{$\text{Ge}^2$-AE~(Ours)}  & IN1K & 48.9 & \textbf{54.1}\\
		\bottomrule
	\end{tabular}
\end{table}
From Tab.~\ref{tab:ade20k}, we can also observe the similar performance gap between MIM-based methods and contrastive learning-based ones. For this dense prediction task, our framework adopting ViT base as encoder can consistently achieve competitive performance than previous methods while adopting larger size ViT large encoder can outperform them evidently.

\textbf{Classification on Other Datasets.}\quad Tab.~\ref{tab:transfer classification} shows 
the transfer learning performance on other classification datasets, including Cifar10~\cite{krizhevsky2009learning}, Cifar100~\cite{krizhevsky2009learning}, iNaturalist18~\cite{van2018inaturalist}, iNaturalist19~\cite{van2018inaturalist}, Flowers~\cite{nilsback2008automated} and Cars~\cite{krause20133d}. For all six datasets, consistent performance improvements can be observed, which verifies great transferability of our $\text{Ge}^2$-AE on classification task.  

\subsection{Ablation Study}

\begin{table*}[htb]
	\small
	\centering
	\caption{Ablation study of $\text{Ge}^2$-AE on IN1K dataset based on the ViT-B architecture. ``PD.'' and ``FD.'' represent pixel and frequency decoders while ``-re'' and ``-con'' denote reconstruction and constraints. ``FT.'' and ``LP.'' are short for fine-tuning and linear probing settings.}
	\label{tab:ablation}
	\begin{tabular}{l|cc|cc|c|c}
		\toprule
		\multirow{2}{*}{Method} & \multicolumn{2}{c|}{Pixel decoder} & \multicolumn{2}{c|}{Frequency decoder} & FT. & LP. \\
		& Pix-re & Freq-con & Freq-re & Pix-con & Top-1(\%) & Top-1(\%)  \\
		\midrule
		$\text{Ge}^2$-AE  w/ Pix. & \cmark& \xmark& \xmark& \xmark&  83.6& 68.0 \\
		$\text{Ge}^2$-AE  w/o FD. & \cmark& \cmark& \xmark& \xmark&  83.8& 68.3 \\
		$\text{Ge}^2$-AE  w/ Freq. & \xmark& \xmark& \cmark& \xmark&  82.2& 53.9 \\
		$\text{Ge}^2$-AE  w/o PD. & \xmark& \xmark& \cmark& \cmark&  82.9& 55.7\\
		$\text{Ge}^2$-AE  w/o con. & \cmark& \xmark& \cmark& \xmark&  84.4& 72.5\\
		\midrule
		\textbf{$\text{Ge}^2$-AE} & \cmark& \cmark& \cmark& \cmark& \textbf{84.8}& \textbf{75.3}\\
		\bottomrule
	\end{tabular}
\end{table*}

To better investigate the effectiveness of different components in our proposed $\text{Ge}^2$-AE, we conduct juxtaposing ablation studies on IN1K dataset under both end-to-end fine-tuning~(FT) and linear probing~(LP) settings, reported in Tab.~\ref{tab:ablation}. In ``$\text{Ge}^2$-AE  w/ Pix.'', our method degraded to the baseline MAE method, where only pixel reconstruction is preserved. By appending frequency constraint to it, as indicated by ``$\text{Ge}^2$-AE  w/o FD.'', both fine-tuning and  and linear probing Top-1 accuracies witness round 0.3\% increase. However, in $\text{Ge}^2$-AE  w/ Freq., when we remove PD and only keep FP preserved but without pixel constraint, there is a 1.4\% drop on fine-tuning performance while linear probing suffers more drastically performance devastation. We attribute it to the ``over-smoothing problem'', in which the model over-concerns the global semantic information while some informative details are abandoned. Although it can be alleviated by appending the pixel constraint~($\text{Ge}^2$-AE w/o PD.), the performance of linear probing setting is still undermined. To further explain this phenomenon, under fine-tuning setting, the contents contained in the lost frequencies~(often the high-frequency) can be hunted back according to F-principle~\cite{xu2019frequency}, whereas no chance for linear probing setting. As a full version, $\text{Ge}^2$-AE can fully leverage the information from both pixel and frequency space, which achieves promising results for both FT and LP settings.

\begin{figure}[t]
	\centering
	\subfloat[Power Law curves]{\includegraphics[width=0.48\linewidth]{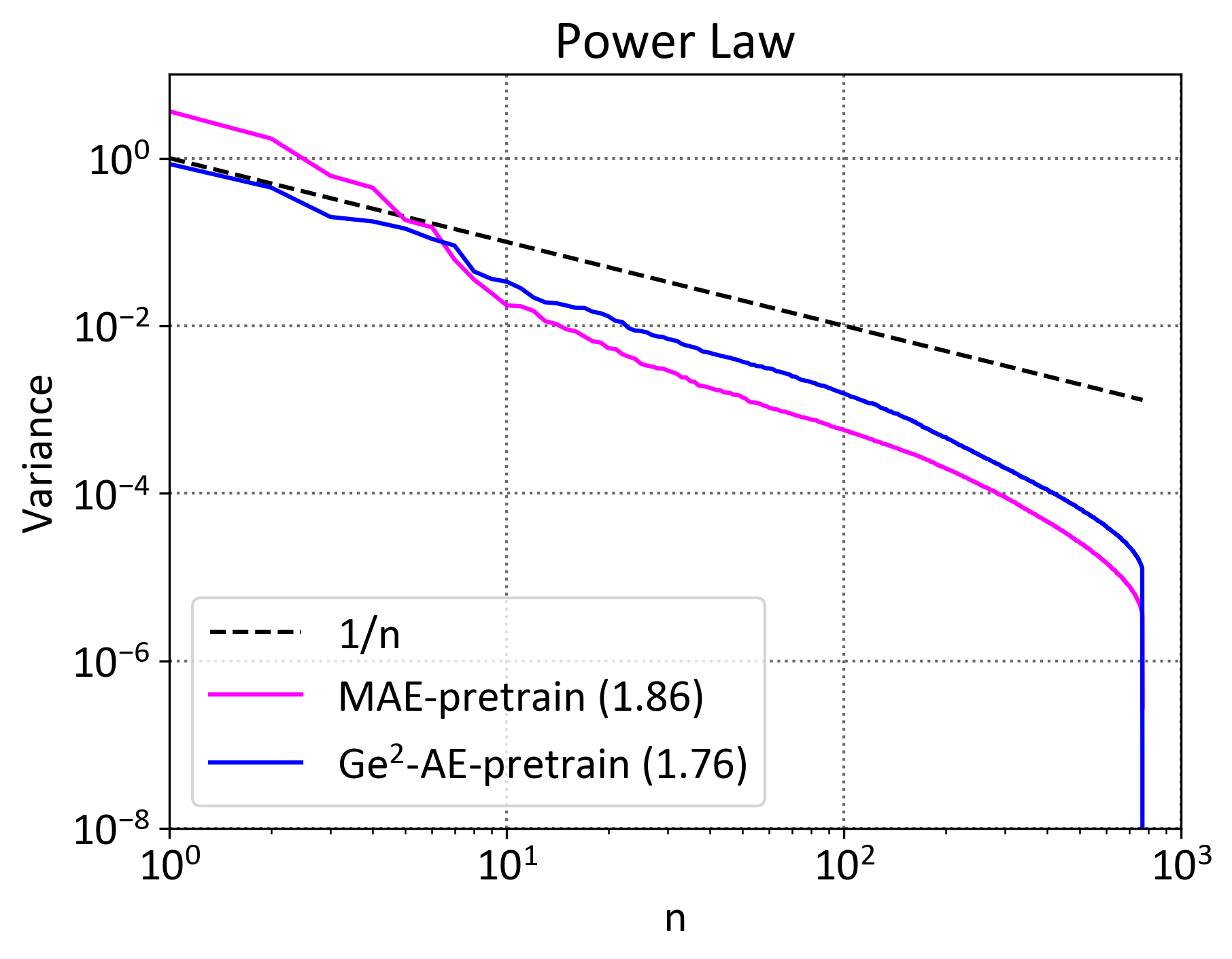}}
	\subfloat[CKA similarity curves]{\includegraphics[width=0.48\linewidth]{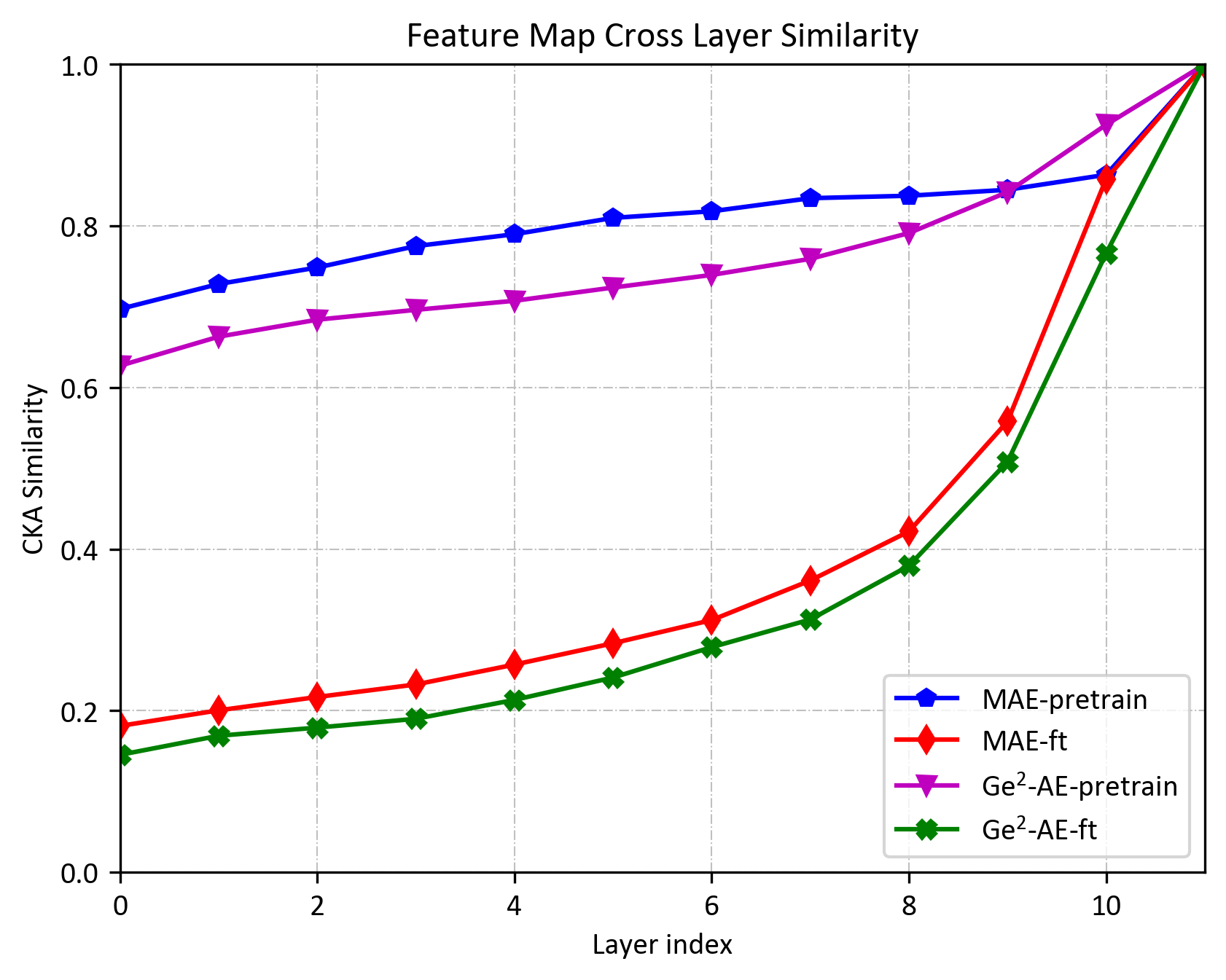}}	
	\caption{Power Law curves and CKA similarity curves of MAE and our $\text{Ge}^2$-AE after pre-trained and fine-tuned. All experiments are performed on IN1K validation set with ViT base model adopted as encoder.} 
	\label{fig:pl}
\end{figure}

\subsection{Analysis on $\textbf{Ge}^2$-AE}\label{ana_geae}
In this subsection, we will further investigate the learning pattern and evaluate the representation quality from both quantitative and qualitative aspects.

\textbf{Power Law Analysis.}\quad According to the recent research in the vision neuroscience~\cite{stringer2019high} and the Machine Learning~\cite{nassar20201} field, the eigenspectrum of the deep feature covariance matrix often follows a Power Law, of which the coefficient $\alpha$ is strongly correlated to the robustness and generalization of representation. Moreover, as another merit, this protocol requires no task-specific labeled data, which is more friendly to self-supervised representation quality evaluation. Therefore, we caculate the eigenspectrum of the empirical feature covariance matrix of the encoder pre-trained by our method and estimate its coefficient $\alpha$, which is defined as:
\begin{align}
	\Sigma_{N}\left(\mathbf{f}_{\theta}\right)=\frac{1}{N} \sum_{i=1}^{N} \mathbf{f}_{\theta}\left(x_{i}\right) \mathbf{f}_{\theta}\left(x_{i}\right)^{\top},
\end{align}
where $\mathbf{f}_{\theta}(\mathbf{x})$ is the representation. After applying spectral decomposition, the eigenvalues $\lambda_{1} \geq \lambda_{2} \cdots \geq \lambda_{n}$ are obtained, which are all nonnegative and follow Power Law:
\begin{align}
	\lambda_{j} \propto j^{-\alpha}.
\end{align}
Here, $\alpha$ is the slope of the Power Law. As suggested by work~\cite{nassar20201}, \textit{$\alpha$ close to 1 indicates the representation is inclined to exhibit robustness and good generalization.} As demonstrated by Power Law curves in Fig.~\ref{fig:pl}(a), the representation pre-trained by our $\text{Ge}^2$-AE have $\alpha$ value (blue curve, 1.76) closer to 1~(corresponding to dotted line) than that of MAE~(magenta curve, 1.86), which confirms the superior quality of the learned representation by our method.

\begin{figure}[t]
	\begin{center}
		\includegraphics[width=0.9\linewidth]{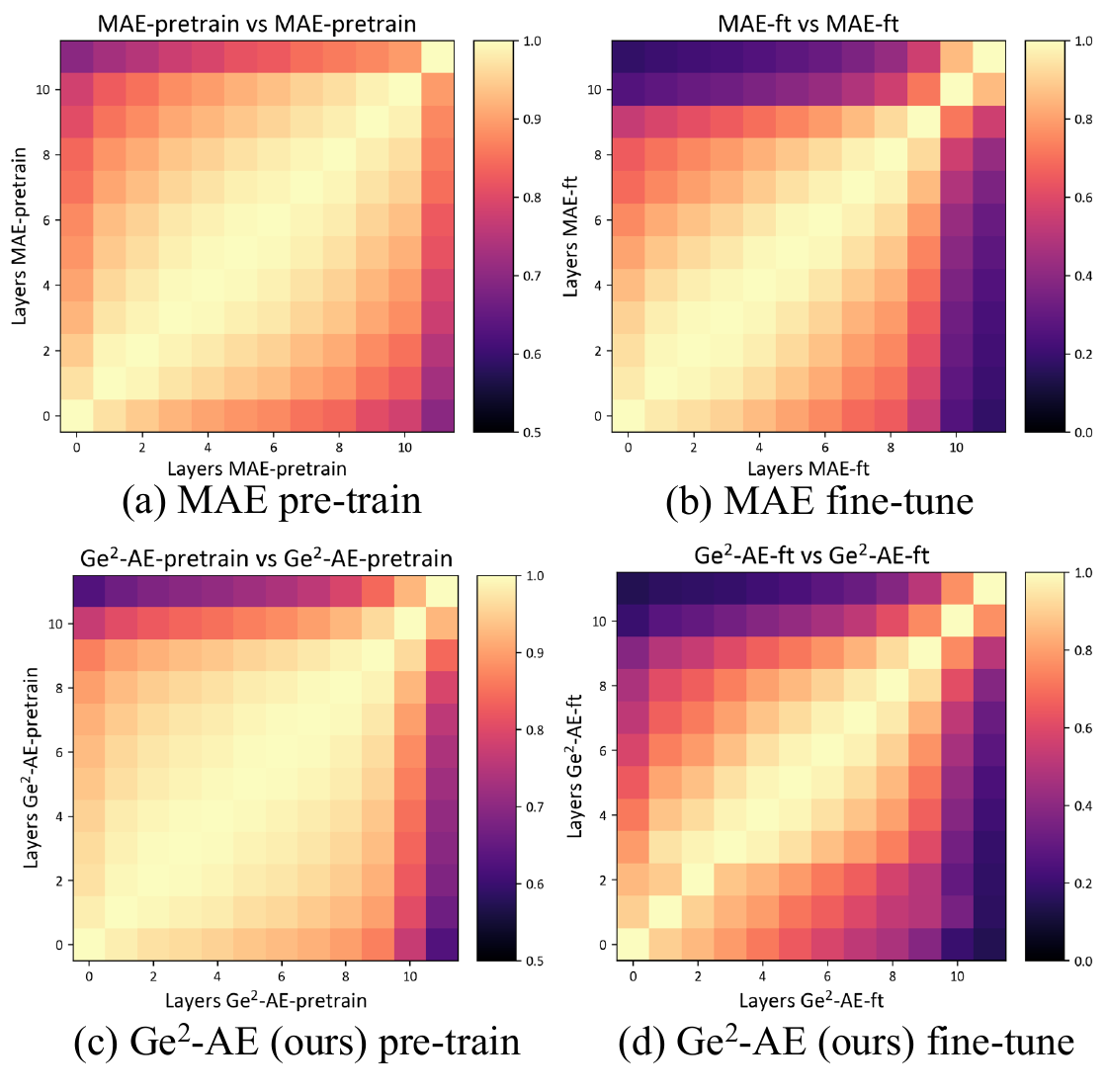}
		{\vspace{-3mm}}
	\end{center}
	\caption{CKA similarities between all pairs of layers across MAE and our $\text{Ge}^2$-AE trained on IN1K. The horizontal axes of heatmaps and vertical axes indexing the layers from input to output. Best viewed in color and zoom in.
	}\label{fig:cka}
	{\vspace{-2mm}}
\end{figure}

\textbf{CKA Similarities of Learned Representation.}\quad The Power Law analysis reveals relationship between the distribution of intermediate features of representation and its robustness. To further study the representation structure learned in our $\text{Ge}^2$-AE, we plot the Centered Kernel Alignment~(CKA) similarities~\cite{kornblith2019similarity} between all pairs of layers across MAE and our $\text{Ge}^2$-AE after pre-trained or fine-tuned in Fig.~\ref{fig:cka}. We can observe that the fine-tuned features (Fig.~\ref{fig:cka}(b) and (d)) often present more compact pattern than pre-trained ones,\textit{i.e.}, merely neighboring several layers can have high similarities when supervision information participating learning. As both MAE and our $\text{Ge}^2$-AE adopting the encoder with same architecture~(ViT-B), the representation structures are somewhat similar, but our method can still exhibit more compact structures~(Fig.~\ref{fig:cka}(c) and (d)) than MAE~(Fig.~\ref{fig:cka}(a) and (b)) for both training settings. 

To profile the ``\textit{evolving speed}'' of learned representations, we also present the CKA similarity curves in Fig.~\ref{fig:pl}(b), where similarities are computed between each layer and the last one. As concluded in previous work~\cite{zhou2021refiner, xu2021evo},if the features change slowly when traversing the model layers, the model could perform inferior to the ones with faster feature evolving speed. By comparison, both pre-trained and fine-tuned features of $\text{Ge}^2$-AE witness faster evolving speed than MAE, which further verifies the effectiveness of our method. 

\begin{figure*}[ht]
	\begin{center}
		\includegraphics[width=0.95\linewidth]{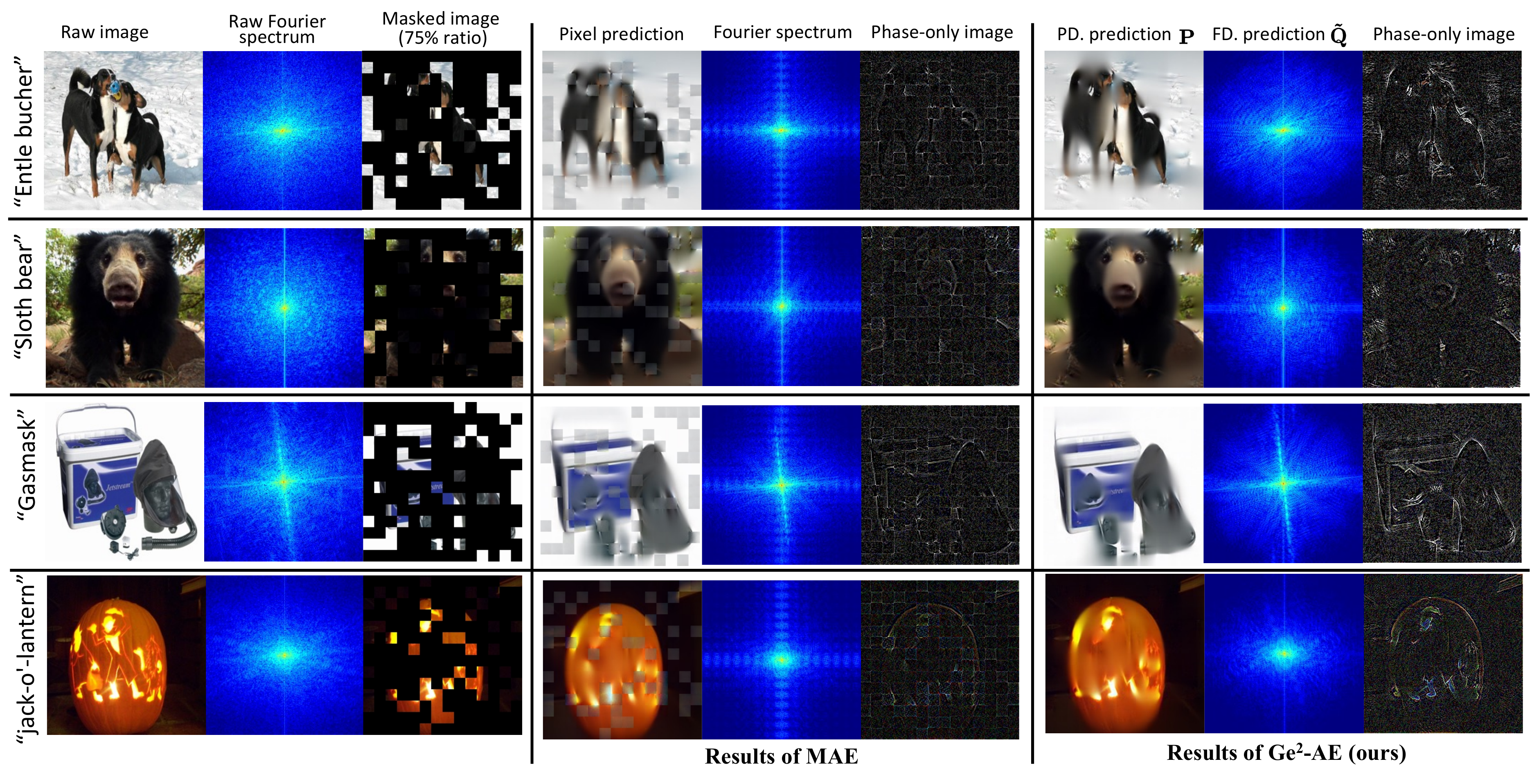}
	\end{center} 
	\caption{Visualizations of the predicted results from MAE and our Geminated Gestalt Autoencoder~($\text{Ge}^2$-AE) pre-trained on IN1K dataset. ``PD.'' and ``FD.'' are short for pixel decoder and frequency decoder respectively. Their outputs are $\mathbf{P}$ and $\mathbf{\tilde{Q}}$, corresponding to the counterparts in Fig.~\ref{fig:arch}(a). Our method can yield results with more necessary global frequency and local details than MAE to overcome the ``over-smoothing'' issue. Best viewed in color and zoomed in.
	}\label{fig:vis}
\end{figure*}

\textbf{Qualitative Analysis}\quad From qualitative aspect, in Fig.~\ref{fig:vis}, we visualize the predicted results of baseline MAE~\cite{he2021masked} and our $\text{Ge}^2$-AE adopting ViT-B as encoder and taking masked images with 75\% ratio. More concretely, the results in ``Fourier spectrum'' column from MAE are obtained by applying 2D-FFT on recovered images in ``pixel prediction'' column, while ``phase-only images'' are obtained by setting the amplitude of Fourier spectrum to constant and restored to pixel image through 2D-IFFT. 
Visually, one can observe that the phase-only images of MAE present corrupted object contour, and the corresponding Fourier spectrum maps may contain some unwanted frequency components while deserved local details are ignored. We suppose this problem is caused by the inherent mechanism in MAE, where only the pixels at masked locations participate the pixel loss computation, while those unmasked ones are also effected uncontrollably. As attempted in MAE, the problem can be solved by calculating loss for all pixels, whereas the performance decrease is witnessed. It is probably because the model is inclined to directly copy the original pixel values of unmasked ones under computing all pixel loss setting while masked pixels are less focused on. 

\begin{figure}[ht]
	\begin{center}
		\includegraphics[width=1\linewidth]{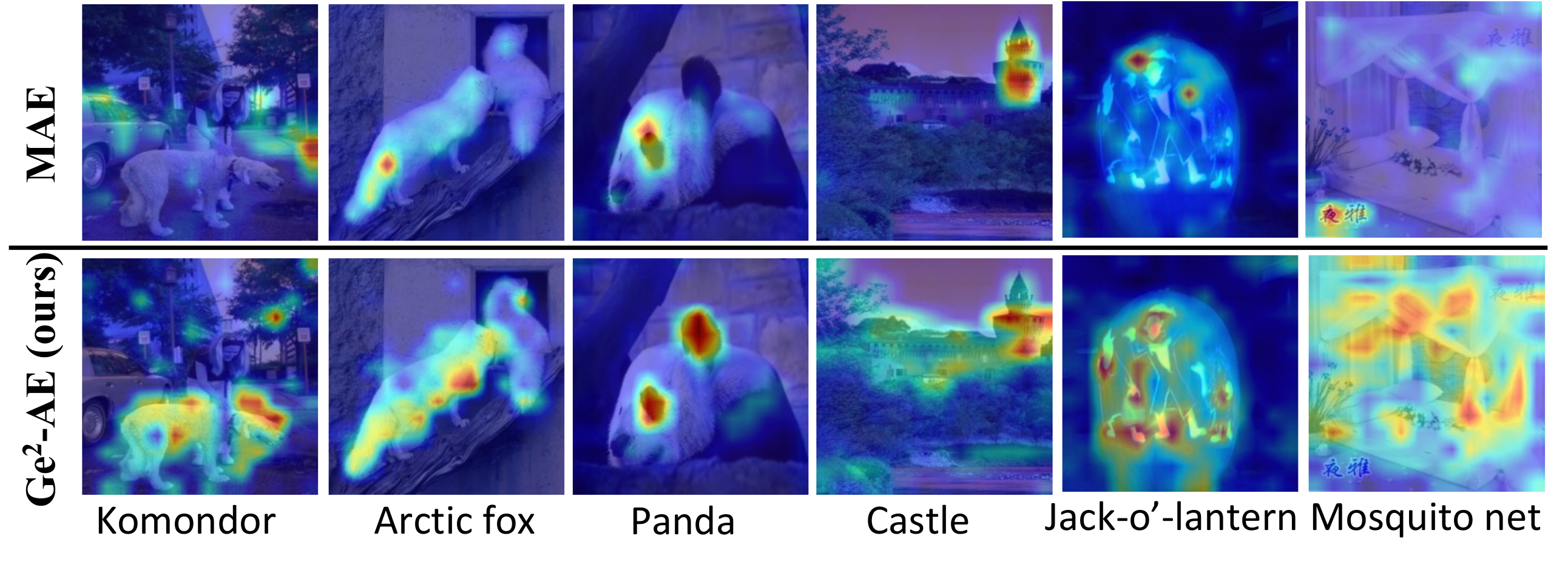}
		{\vspace{-4mm}}
	\end{center}
	\caption{Class activation attention maps on classification task under linear probing setting. Best viewed in color and zoom in.} 
	\label{fig:linear}
\end{figure}

Comparatively, our method simultaneously reconstruct pixel and frequency with mutual constraints. Thanks to this advanced mechanism, in Fig.~\ref{fig:vis}, the predictions of pixel decoder~(``PD. prediction $\mathbf{P}$'') can generate pixel image without unmasked local details lost. For example, the eyes of ``Sloth bear''. Correspondingly, the predictions of frequency decoder can output Fourier spectrum maps $\mathbf{\tilde{Q}}$ with more global frequencies preserved. Consequently, the phase-only images exhibit more reasonable semantic properties with more obvious and holistic object contours. More visualizations are given in appendix.

\textbf{Class Activation Maps of Linear Probing.}\quad Based on interpretability tool~\cite{chefer2021transformer}, in Fig.~\ref{fig:linear}, we also visualize the attention maps of class activation from the model trained under linear probing setting, where pre-trained encoder part is frozen. From the Fig.~\ref{fig:linear}, we find that the attentions of MAE often concentrate on the more local regions, which may caused by the missing of proper global  semantic information. In a stark contrast, our method can present more reasonable attention regions with more scalability. For example, MAE only focuses on the single eye region of ``panda'' while our $\text{Ge}^2$-AE highlights both eye and ear regions with black color, which is more consistent with the exclusive characteristic of panda. Besides, for the ``mosquito net'' case, MAE wrongly pays attention to the logo while the attended region of our method nearly covers the whole object, which is attributable to the superior global semantic abstraction ability brought by our well tailored design. 

\section{Conclusion and Future Work}

We in this paper rethink the Masked Image Modeling~(MIM) for visual pre-training task from frequency perspective and propose a novel $\text{Ge}^2$-AE with dual decoders to reconstruct image contents from both pixel and frequency spaces. Extensive experiments on many downstream visual tasks verify the superior performance of our method than state-of-the-arts. A series of analytic experiments are also conducted to further investigate and explain our model from both 
quantitative and qualitative aspects. We believe shifting perspective to frequency domain could be an enlightening attempt to the community. However, there may still exist redundancy in frequency. We suppose the frequency information could be encoded in more compact format, which can be achieved by introducing more deductive method, such as quantization, in the future.

\begin{appendices}
\renewcommand\thesection{\Alph{section}} 
\renewcommand\thesubsection{\Alph{section}.\arabic{subsection}} 
\renewcommand\thefigure{\Alph{section}\arabic{figure}} 
\renewcommand\thetable{\Alph{section}\arabic{table}} 
\setcounter{section}{0}
\setcounter{figure}{0}	
\setcounter{table}{0}

\section{Centered Kernel Alignment Similarity}

In Sec.~\ref{ana_geae} of main text, we introduce Centered Kernel Alignment (CKA) similarities~\cite{kornblith2019similarity} to evaluate the quality of learned representations. Given the representations from two layers $\mathbf{X}_1 \in \mathbb{R}^{m \times c_1}$ and $\mathbf{X}_2 \in \mathbb{R}^{m \times c_2}$, we elaborate on the detailed definition of CKA similarity as follows:
\begin{align*}
	\textit{CKA}(\mathbf{K}, \mathbf{L})&=\frac{\textit{HSIC}(\mathbf{K}, \mathbf{L})}{\sqrt{\textit{HSIC}(\mathbf{K}, \mathbf{K}) \textit{HSIC}(\mathbf{L}, \mathbf{L})}},\\
	\mathbf{K}& = \mathbf{X}_1\mathbf{X}_1^\top, \mathbf{L} = \mathbf{X}_2\mathbf{X}_2^\top, 
\end{align*} 
where $\textit{HSIC}$ represents the Hilbert-Schmidt Independence Criterion measuring the similarity of centered similarity matrices: 
\begin{align*}
	\textit{HSIC} (\mathbf{K}, \mathbf{L})&= \frac{\textit{vec}(\mathbf{K}^\prime)\cdot \textit{vec}(\mathbf{L}^\prime)}{(m-1)^2},\\
	\mathbf{K}^\prime  &= \mathbf{H}\mathbf{K}\mathbf{H}, \mathbf{L}^\prime  = \mathbf{H}\mathbf{L}\mathbf{H}.
\end{align*}
Specially, \textit{vec}($\cdot$) denotes vectorization operation. $\mathbf{H}=\mathbf{I}_{m}-\frac{1}{m} \mathbf{11}^\top$ is the centering matrix.

\newlength\savewidth\newcommand\shline{\noalign{\global\savewidth\arrayrulewidth
		\global\arrayrulewidth 1pt}\hline\noalign{\global\arrayrulewidth\savewidth}}
\newcommand{\tablestyle}[2]{\setlength{\tabcolsep}{#1}\renewcommand{\arraystretch}{#2}\centering\footnotesize}
\newcolumntype{x}[1]{>{\centering\arraybackslash}p{#1pt}}
\newcolumntype{y}[1]{>{\raggedright\arraybackslash}p{#1pt}}
\newcolumntype{z}[1]{>{\raggedleft\arraybackslash}p{#1pt}}

\section{Detailed Training Recipes}
\subsection{ImageNet Experiments}
\noindent\textbf{Pre-training.} The detailed default setting is given in Tab.~\ref{tab:supp_pre}, with linear \textit{lr} scaling rule and \textit{xavier\_uniform} initialization applied~\cite{he2021masked}. 

\noindent\textbf{End-to-end fine-tuning and linear probing.}
Our fine-tuning and linear classifier training also inherit the training recipes from MAE~\cite{he2021masked}, which are elaborated in Tab.~\ref{tab:supp_finetune} and Tab.~\ref{tab:supp_linear}. 

\subsection{Transfer Learning Experiments}
For the \textbf{Objection detection on COCO} and \textbf{Semantic segmentation on ADE20K} tasks, the detailed settings are related in the Tab.~\ref{tab:supp_det} and Tab.~\ref{tab:supp_seg}. As for the classification on other six datasets, we follow the training recipe described in Tab.~\ref{tab:supp_other}.  

\begin{table}[t]
	\tablestyle{6pt}{1.02}
	\caption{Pre-training setting.}\label{tab:supp_pre}
	\begin{tabular}{y{96}|y{68}}
		\toprule
		Config & Value \\
		\midrule
		optimizer & AdamW \cite{loshchilov2017decoupled} \\
		base learning rate & 1.5e-4 \\
		weight decay & 0.05 \\
		optimizer momentum & $\beta_1, \beta_2{=}0.9, 0.95$ \cite{chen2020generative} \\
		batch size & 4096 \\
		learning rate schedule & cosine decay \cite{loshchilov2016sgdr} \\
		warmup epochs \cite{goyal2017accurate} & 40 \\
		augmentation & RandomResizedCrop \\
		\bottomrule
	\end{tabular}
\end{table}

\begin{table}[t]
	\tablestyle{6pt}{1.02}
	\caption{End-to-end fine-tuning setting.}\label{tab:supp_finetune}
	\begin{tabular}{y{96}|y{68}}
		\toprule
		Config & Value \\
		\midrule
		optimizer & AdamW~\cite{loshchilov2017decoupled} \\
		base learning rate & 1e-3 \\
		weight decay & 0.05 \\
		optimizer momentum & $\beta_1, \beta_2{=}0.9, 0.999$ \\
		layer-wise lr decay \cite{clark2020electra,bao2021beit} & 0.75 \\
		batch size & 1024 \\
		learning rate schedule & cosine decay \\
		warmup epochs & 5 \\
		training epochs & 100 (B), 50 (L/H) \\
		augmentation & RandAug (9, 0.5) \cite{cubuk2020randaugment} \\
		label smoothing \cite{szegedy2016rethinking} & 0.1 \\
		mixup \cite{zhang2017mixup} & 0.8 \\
		cutmix \cite{yun2019cutmix} & 1.0 \\
		drop path \cite{huang2016deep} & 0.1 (B/L) 0.2 (H) \\
		\bottomrule
	\end{tabular}
\end{table}

\begin{table}[t]
	\tablestyle{6pt}{1.02}
	\caption{Linear probing setting.}\label{tab:supp_linear}
	\begin{tabular}{y{96}|y{68}}
		\toprule
		Config & Value \\
		\midrule
		optimizer & LARS~\cite{you2017large} \\
		base learning rate & 0.1 \\
		weight decay & 0 \\
		optimizer momentum & 0.9 \\
		batch size & 16384 \\
		learning rate schedule & cosine decay \\
		warmup epochs & 10 \\
		training epochs & 90 \\
		augmentation & RandomResizedCrop \\
		\bottomrule
	\end{tabular}
\end{table}

\begin{table}[t]
	\tablestyle{6pt}{1.02}
	\caption{Objection detection setting.}\label{tab:supp_det}
	\begin{tabular}{y{96}|y{68}}
		\toprule
		Config & Value \\
		\midrule
		optimizer & AdamW~\cite{loshchilov2017decoupled} \\
		base learning rate & 1.6e-4 \\
		weight decay & 0.1 \\
		optimizer momentum & $\beta_1, \beta_2{=}0.9, 0.999$ \\
		batch size & 64 \\
		learning rate schedule & cosine decay \\
		warmup epochs & 0.25 \\
		training epochs & 100 \\
		augmentation & Large-scale Jitter \\
		\bottomrule
	\end{tabular}
\end{table}

\begin{table}[t]
	\tablestyle{6pt}{1.02}
	\caption{Semantic segmentation setting.}
		\label{tab:supp_seg}
	\begin{tabular}{y{96}|y{68}}
		\toprule
		Config & Value \\
		\midrule
		optimizer & AdamW \cite{loshchilov2017decoupled} \\
		base learning rate & 1e-3 \\
		weight decay & 0.05 \\
		optimizer momentum & $\beta_1, \beta_2{=}0.9, 0.999$ \\
		batch size & 16 \\
		learning rate schedule & linear \\
		warmup steps & 1500 \\
		training steps & 160K \\
		\bottomrule
	\end{tabular}
\end{table}

\begin{table}[t]
	\tablestyle{6pt}{1.02}
	\caption{Other classification settings.}\label{tab:supp_other}
	\begin{tabular}{y{96}|y{68}}
		\toprule
		Config & Value \\
		\midrule
		optimizer & AdamW \\
		base learning rate & 1e-3 \\
		weight decay & 0.05 \\
		optimizer momentum & $\beta_1, \beta_2{=}0.9, 0.999$ \\
		layer-wise lr decay \cite{clark2020electra,bao2021beit} & 0.75 \\
		batch size & 1024 \\
		learning rate schedule & cosine decay \\
		warmup epochs & 5 \\
		training epochs & 100\\
		augmentation & RandAug (9, 0.5) \cite{cubuk2020randaugment} \\
		label smoothing \cite{szegedy2016rethinking} & 0.1 \\
		mixup \cite{zhang2017mixup} & 0.8 \\
		cutmix \cite{yun2019cutmix} & 1.0 \\
		drop path \cite{huang2016deep} & 0.1 \\
		\bottomrule
	\end{tabular}
\end{table}

\section{Effects of Configuration Choice}
\subsection{ Decoder Depth}
\begin{table}
	\small
	\setlength{\tabcolsep}{1mm}
	\caption{Various decoder block numbers to the performance of our  $\text{Ge}^2$-AE~(\textit{ViT-B}) on IN1K dataset.}\label{tab:supp_d_blk}                           
	\begin{tabular}{c|c|c}
		\toprule
		Blocks & Fine-tuning~(top-1 in \%) & Linear probing~(top-1 in \%) \\
		\midrule
		1 & 84.7 & 66.7 \\
		4 & 84.8 & 71.7 \\
		\textbf{8} & \textbf{84.8} & \textbf{75.3} \\
		12 & 84.5 & 75.0 \\
		\bottomrule
	\end{tabular}
\end{table}

As stated in main text, decoders for reconstruction are crucial modules determining the pre-trained representation quality, we thus in Tab.~\ref{tab:supp_d_blk} show the effects of different decoder block numbers on the performance. More concretely, we progressively increase the depth of both decoders~(pixel and frequency), ranging from 1 to 12. Through Tab.~\ref{tab:supp_d_blk}, we find the effect of decoder depth on the ``fine-tuning'' performance is limited, where the fluctuation is less than 0.4\%. Comparatively, the ``linear probing'' setting witnesses more performance variation, which consistently increases with deeper decoder depth, while the performance becomes saturated when block number increased to 8.

\subsection{Importance of Frequency}
In Tab.~\ref{tab:supp_lbd}, we further investigate the importance of the frequency branch in our geminated-structure method. Specifically, the frequency loss weight  $\lambda$ defined in Eqn.~\eqref{overall-loss} of main text is tuned from 0.2 to 2.0. Compared with the decoder depth, the weight change has less effect to both settings and the best performance is observed when $\lambda$ is set to 0.5. It is worth pointing out that if the $\lambda$ is set to big value~(2.0), the performance would drop to certain extent.

\begin{table}
	\small
	\setlength{\tabcolsep}{1.5mm}
	\caption{Various frequency loss weights $\lambda$ to the performance of our  $\text{Ge}^2$-AE~(\textit{ViT-B}) on IN1K dataset.}\label{tab:supp_lbd}                           
	\begin{tabular}{c|c|c}
		\toprule
		$\lambda$ & Fine-tuning~(top-1 in \%) & Linear probing~(top-1 in \%) \\
		\midrule
		0.2 & 84.6 & 74.3 \\
		\textbf{0.5} & \textbf{84.8} & \textbf{75.3} \\
		1.0 & {84.7} & {74.1} \\
		2.0 & 84.5 & 73.4 \\
		\bottomrule
	\end{tabular}
\end{table}

\subsection{Masking Ratio}
\begin{figure}[t]
	\centering
	\includegraphics[width=0.9\linewidth]{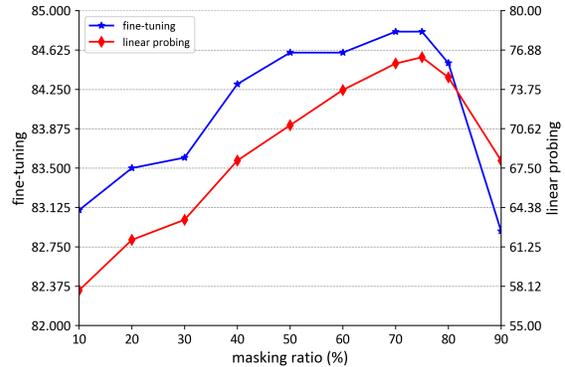}
	\caption{Effects of various mask ratios in pre-training phase to the performance of downstream fine-tuning and linear probing on IN1K dataset with ViT-B as encoder. } 
	\label{fig:supp_mr}
\end{figure}

The influence of masking ratio is shown in Fig.~\ref{fig:supp_mr}, which directly determines the difficulty of reconstruction task. The best performance can be achieved for both fine-tuning and linear probing settings when 75\% masking ratio adopted, which follows similar trends with MAE. To further explain, a large masking ratio is also of advantage to our frequency prediction involved method for learning more informative representation. In addition, ``linear probing'' performance is more sensitive to various masking ratios, than ``fine-tuning''.

\begin{figure*}[ht]
	\centering
	\includegraphics[width=1\linewidth]{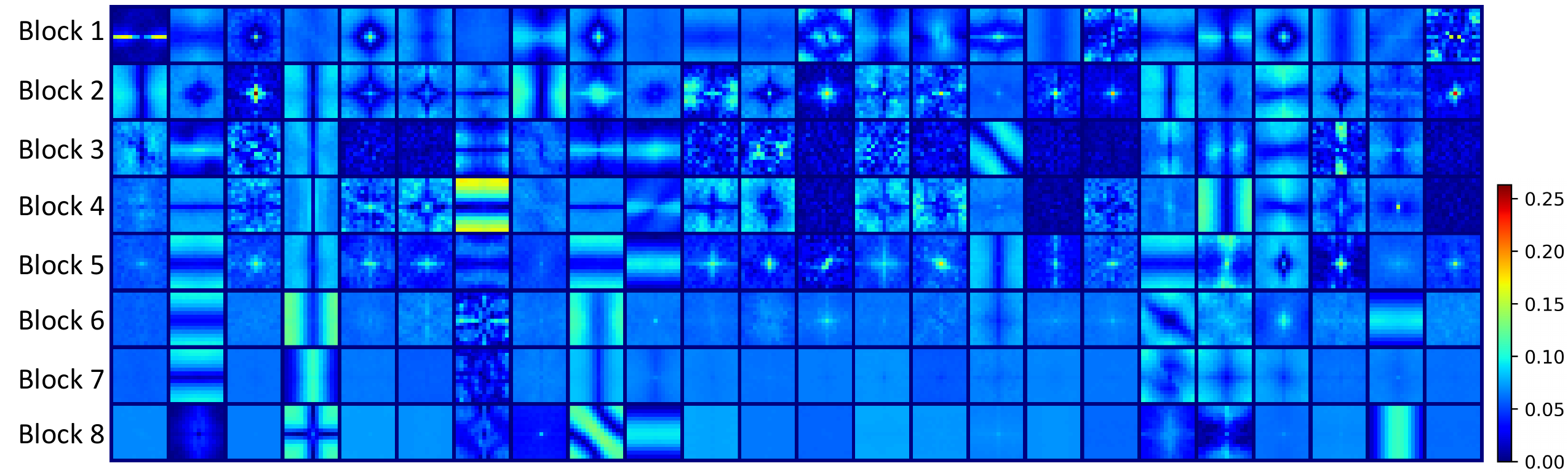}
	\caption{Weight visualization of Fourier Spectrum Perceiver~(FSP) in frequency decoder of $\text{Ge}^2$-AE. Best viewed in color.} 
	\label{fig:supp_fsp}
\end{figure*}

\section{More Qualitative Analysis}

\subsection{Visualization of FSP Weights.}
To explore the learning pattern of Frequency Decoder~(FD) of our $\text{Ge}^2$-AE, we in Fig.~\ref*{fig:supp_fsp} visualize the first 24-channel weights $\Omega$~(Sec.~\ref{sec:fq} in main text) in Fourier Spectrum Perceiver~(FSP) of each FD block. In the first several FD blocks nearing encoder output, FSP weights exhibit diversified patterns, \textit{i.e.,} different channels of FSP emphasize different frequency components where each one corresponds to an exclusive global informative content. Along with block number increasing, the patterns of FSP weights become reasonably sparse, because only those useful frequencies are preserved for reconstruction task during this procedure of global high semantic extraction. Thanks to the rich patterns serving as candidates, more effective information hidden in the frequency can be leveraged. Conclusively, we believe this ``from-dense-to-sparse'' pattern evolution is one of the keys to solve ``over-smoothing'' problem stated in main text.  

\subsection{Class Activation Attention Maps}
\begin{figure*}[ht]
	\centering
	\includegraphics[width=1\linewidth]{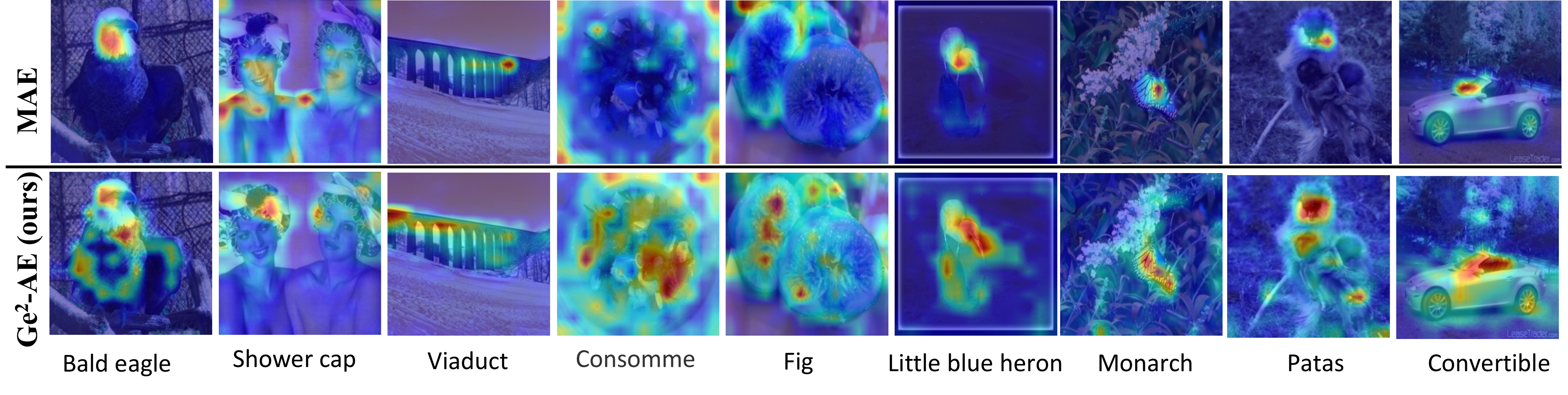}
	\caption{Class activation attention maps on IN1K dataset under linear probing setting.} 
	\label{fig:supp_linear}
\end{figure*}

\begin{figure*}[t]
	\centering	
	\includegraphics[width=0.95\linewidth]{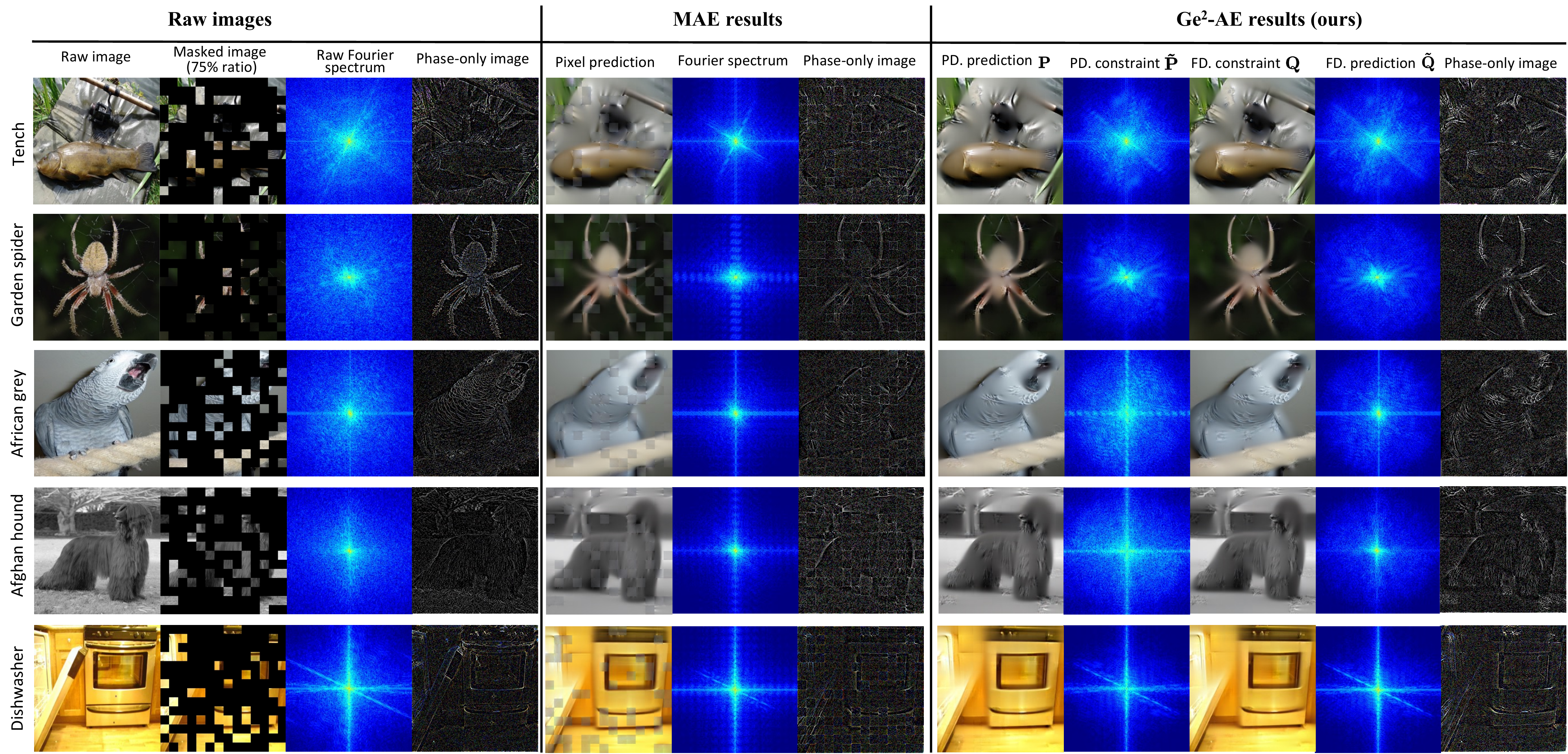}
	\caption{Uncurated random samples of the predicted results from MAE and our Geminated Gestalt Autoencoder~($\text{Ge}^2$-AE) pre-trained on IN1K dataset.} 
	\label{fig:supp_vis}
\end{figure*}
In Fig.~\ref{fig:supp_linear}, we visualize more persuasive evidences, \textit{i.e.}, class activation attention maps of uncurated random samples from IN1K dataset using method~\cite{chefer2021transformer}. As claimed in the main text, ``\textit{a good latent representation should be a consensus between
	local details and global semantics}''. Compared with MAE~\cite{he2021masked}, our $\text{Ge}^2$-AE exhibits more reasonable activation attention maps. For example, in ``consomme'' case, MAE wrongly allocates the attention to the background due to the lack of global perspective while our method can successfully attend to the actual ingredients of the class. Analogously, the similar consequence can also be observed 
on the ``fig'' case where representative pulp texture is highlighted, which also serves as a strong proof to reflect the quality of our method learned representations. Besides, for the ``convertible'' case, we surprisingly find the activation of our $\text{Ge}^2$-AE can focus on the opened roof of the car, which presents the high class-related semantic property.  
\subsection{Visualization of Predicted Results}
To better investigate the learning behavior of our model, we visualize more yielded results of random samples from IN1K in the Fig.~\ref{fig:supp_vis}. Specially, other than the direct predictions from pixel decoder~(PD.) $\mathbf{P}$ and frequency decoder~(FD.)  $\mathbf{\tilde{Q}}$ as in Fig.~\ref{fig:vis} of main text, we also visualize their indirect counterpart constraints obtained by FFT and IFFT, \textit{i.e.}, $\mathbf{\tilde{P}}$ and $\mathbf{Q}$. From Fig.~\ref{fig:supp_vis}, we can observe that, no mater the direct or indirect results of our method, the predicted results from both pixel and frequency space can reach harmony. That is to say, the global frequency can feasibly recover pixels containing details and vice versa, which can also be confirmed by the phase-only images presenting more wholistic contours and more object-related details than that of MAE.

\section{Limitations}
As discussed in Tab.~\ref{tab:knn&linear} of main text, in terms of the ``linear probing'' experimental results on IN1K classification dataset, our proposed method can beat other contemporary MIM-based methods, such as MAE, SimMIM and CAE, by at most 7\%. However, we still observe the performance gap between complex contrastive learning-based methods and ours. We attribute this gap to the ``linear probing'' setting \textit{per se}, as also discussed in the work~\cite{fang2022corrupted}. The contrastive learning-based methods only focus on learning representations from limited 1,000 classes, which is naturally advantageous under this evaluation setting. Nevertheless, every sword has two edges, MIM-based methods often present more generalization ability as what they care about are more than limited classes, which can be demonstrated by the results on other transfer learning tasks~(Tab.~\ref{tab:coco}$\sim$\ref{tab:ade20k} in main text). Despite the gap in ``linear probing'' performance, in contrast to other MIM-based methods, this gap can be largely remedied by our method. We believe there would exist other more objective and neutral protocols to evaluate the representation qualities. But, it may a little bit deviate from the research purpose in this paper and we leave it to our future study.

\section{Broader Impact}
The multimedia content can be presented in various formats or modalities. In this paper, we propose a self-supervised learning method to learn robust representation from unlabeled data in spatial-frequency space. Essentially, Fourier frequency spectrum can be deemed as another modality of visual data. From the multi-modality perspective, our method thus can be interpreted as a novel multi-modality pre-training algorithm mining complementary information from both modalities by unprecedentedly regarding them as reconstruction targets, which is of difference from most previous MIM methods taking single visual data as input and prediction target. Surprisingly, we find our simple yet effective method can successfully propagate both modality information from reconstruction targets to the learned representation. For the community, we provide a new perspective to rethink representation learning task. The frequency decoder~(FD) in our method is a lightweight and play-and-plug module, therefore we encourage researchers to build new pre-training models based on our spatial-frequency adaptable FD we can expect to be particularly beneficial.     
\end{appendices}

{\small
\bibliographystyle{ieee_fullname}
\bibliography{mimbib}
}
\end{document}